\documentclass[letterpaper,10pt,journal]{IEEEtran}

\usepackage{amsmath, amssymb, amsfonts}

\usepackage{graphicx}
\usepackage[Export]{adjustbox}
\usepackage{caption}
\usepackage{subcaption} 
\usepackage{booktabs}
\usepackage{tabularx}
\usepackage{cellspace}
\usepackage[table]{xcolor} 

\usepackage{algorithm}
\usepackage{algpseudocode} 

\usepackage{tikz}
\usepackage{pgfplots}
\pgfplotsset{compat=1.18}
\usepgfplotslibrary{groupplots}

\usepackage{listings}
\usepackage{lipsum}
\usepackage{cite}

\usepackage[hidelinks]{hyperref} 
\usepackage{cleveref} 

\usepackage{xcolor}
\definecolor{myblue}{RGB}{0, 0, 0}
\usepackage{pifont}
\definecolor{colorUnique}{RGB}{78, 154, 204}
\definecolor{colorMultiple}{RGB}{255, 168, 92}
\definecolor{colorVLM}{RGB}{98, 196, 98}
\definecolor{colorNoVLM}{RGB}{82, 128, 188}
\definecolor{colorSky}{RGB}{142, 205, 242}
\definecolor{colorOrangeLight}{RGB}{255, 210, 158}
\definecolor{colorNoVLMSoft}{RGB}{169, 191, 221}

\definecolor{fail1}{RGB}{62, 148, 206}
\definecolor{fail2}{RGB}{244, 188, 76}
\definecolor{fail3}{RGB}{64, 184, 154}
\definecolor{fail4}{RGB}{232, 138, 74}
\definecolor{fail5}{RGB}{132, 198, 238}
\definecolor{fail6}{RGB}{236, 210, 95}
\definecolor{fail7}{RGB}{228, 168, 198}
\definecolor{fail8}{RGB}{58, 132, 158}
\definecolor{fail9}{RGB}{244, 200, 88}
\definecolor{plotLine}{RGB}{186, 196, 210}
\definecolor{plotGrid}{RGB}{232, 238, 245}
\definecolor{plotLegendFill}{RGB}{248, 250, 252}
\newsavebox{\arrangebox}

\DeclareMathOperator*{\argmax}{arg\,max}

\title{\LARGE \bf
SceneGraphGrounder: Zero-Shot 3D Visual Grounding via Structured Scene Graph Matching
}

\author{Xuefei Sun \quad Xujia Zhang \quad Brendan Crowe \quad Doncey Albin \quad Christoffer Heckman
\thanks{The authors are with University of Colorado Boulder. {\tt\small \{firstname.lastname\}@colorado.edu}}
\thanks{This work was sponsored by the Army Research Laboratory under Cooperative Agreement Number W911NF-17-2-0181.}
\thanks{This work is under review for IEEE RAL.}
}

\begin{document}

\maketitle
\thispagestyle{empty}
\pagestyle{empty}

\begin{abstract}
Zero-shot 3D visual grounding requires localizing objects in unstructured environments from free-form natural language. Recent vision–language model (VLM) approaches achieve promising results but rely on view-dependent reasoning or implicit representations, limiting spatial consistency and interpretability for compositional queries. We propose SceneGraphGrounder, a framework that reformulates 3D grounding as structured graph matching over a reconstructed 3D scene graph.  To enable this formulation, we introduce a visual marker prompting strategy that enables a VLM to infer object–object relationships from 2D views, which are subsequently lifted into a persistent 3D scene graph encoding both spatial and semantic relations. Given a query, we construct a query graph and perform constrained alignment with the scene graph, ensuring multi-view consistency and interpretable reasoning. Experiments on the ScanRefer benchmark demonstrate that our method achieves competitive performance among zero-shot approaches, using only RGB-D inputs. We further validate our framework through real-world deployment on a mobile robot, demonstrating robust spatial reasoning in long-horizon physical environments. We will make our code publicly available upon acceptance.
\end{abstract}    
\section{Introduction}
\label{sec:intro}

3D visual grounding refers to localizing a target object in a 3D scene using a free-form natural language description provided by a human. The task becomes particularly challenging in zero-shot and open-vocabulary settings, where object categories and spatial relations are not constrained to predefined labels. In these scenarios, models must interpret diverse object references, compositional descriptions, and nuanced spatial cues, demanding a deeper and more flexible form of semantic understanding beyond fixed label spaces.

Recent approaches to address these challenges have increasingly turned to large vision-language models (VLMs) and large language models (LLMs). In particular, agent-based frameworks formulate grounding as an iterative reasoning process over multi-view observations or external tools~\cite{yang2023llmgrounderopenvocabulary3dvisual,xu2024vlmgroundervlmagentzeroshot}. 
In parallel, alternative methods seek to bridge 3D scenes with 2D VLMs via rendering and viewpoint adaptation, enabling powerful image-based models to operate on 3D data~\cite{li2025seegroundgroundzeroshotopenvocabulary,yuan2024visualprogrammingzeroshotopenvocabulary}. While these approaches have achieved strong performance, they largely rely on implicit scene representations, reasoning directly over rendered views, proposal features, or intermediate tool outputs. As a result, spatial relationships are inferred on-the-fly rather than explicitly stored, leading to potential inconsistencies across viewpoints and requiring repeated reasoning for each query, which limits efficiency and prevents reuse of scene structure. The lack of an explicit relational representation further reduces interpretability and weakens robustness on compositional queries involving multiple objects and spatial constraints.

{\color{myblue}
These limitations raise a fundamental question: \textit{how should a 3D environment be represented for grounding and language reasoning?} Prior work explores a spectrum of representations, from CLIP-aligned semantic maps\cite{kerr2023lerf,Peng2023OpenScene} to 3D hierarchical scene graph~\cite{ray2024taskmotionplanninghierarchical}. At the same time, natural language descriptions are inherently relational and compositional, and they can be naturally represented as graphs~\cite{bonn2020spatial}, where entities correspond to nodes and semantic relations define edges. Similarly, in the robotics field, grounding language can also be formulated as graphs~\cite{paul2016efficient}, enabling compositional interpretation through graph-based alignment between linguistic structure and the world. 

This shared graph-like structure of both language queries and 3D scenes motivates our approach: by explicitly constructing a graph for the scene and aligning it with a graph derived from the language query, grounding can be reformulated as a structured matching problem rather than relying on repeated implicit reasoning at inference time.}

In this work, we present \textbf{SceneGraphGrounder}, a framework that formulates 3D visual grounding as structured graph matching between a query graph and a 3D scene graph. We represent queries as graphs and perform grounding via graph matching to a semantically rich 3D scene graph constructed from RGB-D data, augmented with VLM-inferred object relationships transferred from 2D views. 

{
\color{myblue}
The primary contributions of this work are:
\begin{itemize}
\item Formulate 3D visual grounding as a structured graph matching problem between a query graph and a reconstructed 3D scene graph, improving efficiency and interpretability.
\item Construct a semantically rich 3D scene graph that encodes both spatial and semantic relationships, enabling structured and reusable relational reasoning, to support graph matching. 
\item Enable zero-shot 3D visual grounding from raw RGB-D observations and odometry, without ground-truth point clouds or precomputed object proposals. Our work achieves competitive performance with prior zero-shot methods that rely on privileged 3D inputs.
\end{itemize}
}

\section{Related Works}
\label{sec:RW} 

\begin{figure*}[t]  
\centering
\includegraphics[width=0.95\textwidth]{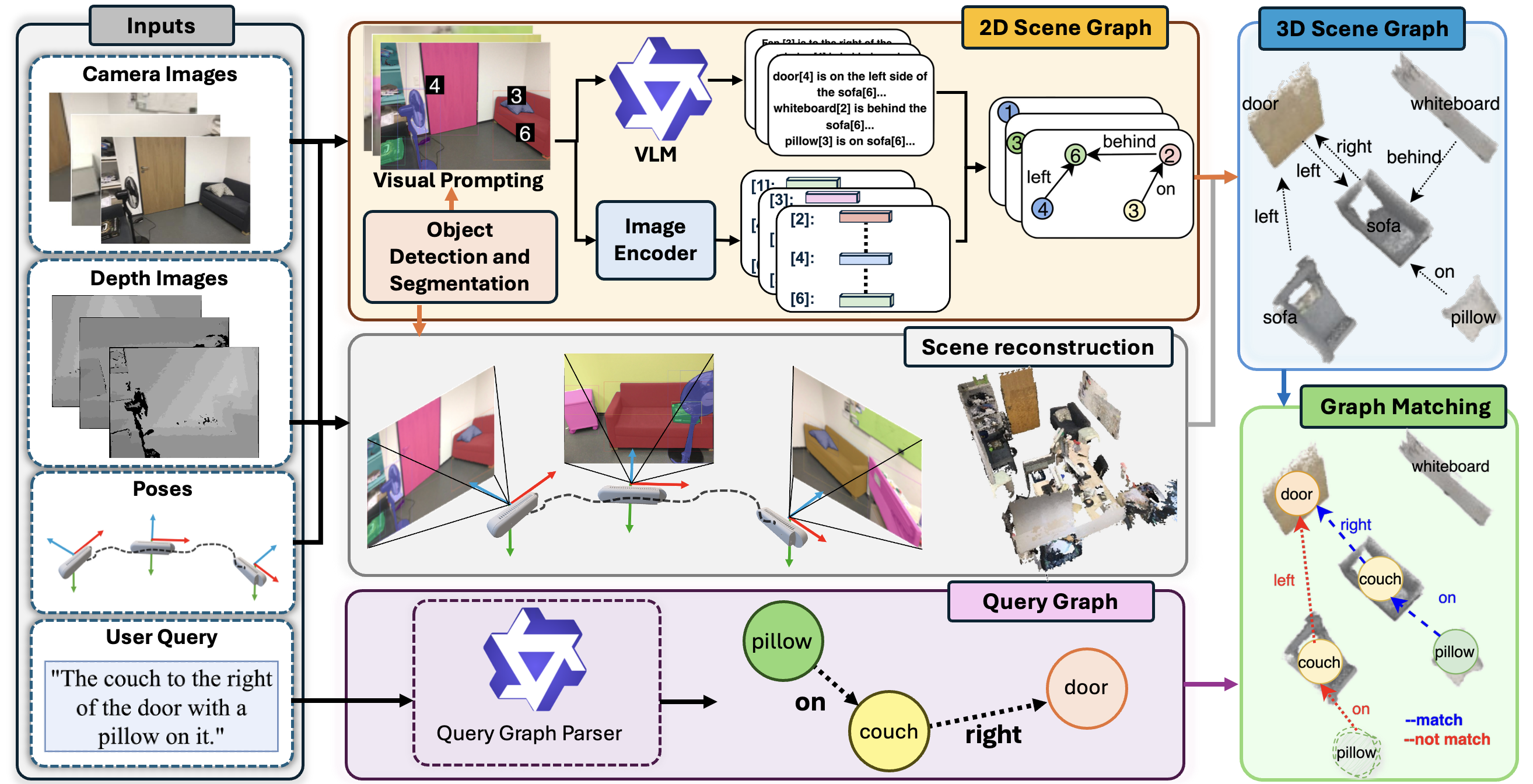}
\caption{\textbf{System Overview.} Our framework takes RGB-D images, sensor odometry, and a user query as input. The graph is lifted into 3D via visual prompting and association with the reconstructed point cloud, forming a global 3D scene graph. Object grounding is achieved via graph matching with the parsed query.}
\label{fig:system_pipeline}
\end{figure*}

\subsection{Scene Graph Generation}
2D scene graph generation (SGG) has increasingly shifted toward VLM-based formulations, where the task is reframed as multimodal reasoning or generation over object and relationship \cite{xu2024llavaspacesggvisualinstructtuning,yu2023visuallypromptedlanguagemodelfinegrained}. Recent open-vocabulary approaches further improve generalization to unseen objects and predicates by leveraging large VLMs and language priors. Within this line of work, early methods primarily rely on prompt-based designs, whereas more recent approaches shift toward zero-shot alignment \cite{chen2024expanding,xu2024llavaspacesggvisualinstructtuning}, probabilistic grounding \cite{liu2025relationawarehierarchicalpromptopenvocabulary}, and distillation strategies \cite{chen2024scenegraphgenerationroleplaying} built on pretrained models such as CLIP~
\cite{radford2021learningtransferablevisualmodels}.

Recent research has extended SGG beyond 2D to 3D environments using point clouds or multi-view RGB images, broadly categorized into methods leveraging explicit 3D geometry for contextual reasoning \cite{wald2020learning3dsemanticscene,wang2023vlsatvisuallinguisticsemanticsassisted,zhang2021exploitingedgeorientedreasoning3d} and those operating on multi-view images, sometimes enhanced with depth or SLAM-based reconstruction \cite{wu2021scenegraphfusionincremental3dscene,wu2023incremental3dsemanticscene}. Prior work explores pairwise relationship modeling \cite{wald2020learning3dsemanticscene}, GCN-based aggregation and edge-centric reasoning \cite{wu2021scenegraphfusionincremental3dscene,yang2024exploiting}, as well as integration of commonsense knowledge and external priors \cite{NEURIPS2021_9a555403,feng20233d}, or knowledge distillation from multimodal teachers \cite{wang2023vlsatvisuallinguisticsemanticsassisted}. Despite these advances, most approaches still focus on constructing a global scene graph and struggle to incorporate structured, task-driven reasoning.

\subsection{3D Visual Grounding}
3D visual grounding aims to localize objects in 3D scenes based on natural language descriptions. Early methods \cite{chen2020scanrefer3dobjectlocalization,achlioptas2020referit3d} relied on supervised learning with aligned 3D annotations and fixed object categories, typically using point cloud encoders or proposal-based frameworks. While effective in closed-set settings, their generalization to unseen categories and compositional expressions is limited. Recent open-vocabulary approaches leverage vision–language representations to align text with 3D object proposals\cite{3dvista,yuan2021instancerefer,Peng2023OpenScene}, but they often rely on object-level similarity and lack explicit relational reasoning, limiting robustness for complex expressions. Recent transformer-based approaches improve global multimodal fusion, and this line of work is further extended by methods such as 
~\cite{li2025seegroundgroundzeroshotopenvocabulary}, which leverages multi-view reasoning and VLM integration to achieve strong zero-shot and open-vocabulary performance. Building on similar ideas, agent-based frameworks~\cite{yang2023llmgrounderopenvocabulary3dvisual,xu2024vlmgroundervlmagentzeroshot} decompose language queries and coordinate perception tools or multi-view reasoning. In a related direction, methods have been proposed that bridge 3D scenes with 2D VLMs through rendering-based and modular reasoning strategies~\cite{li2025seegroundgroundzeroshotopenvocabulary, yuan2024visualprogrammingzeroshotopenvocabulary}.
Despite their effectiveness, these approaches typically rely on implicit scene representations and perform relational reasoning dynamically during inference, making them sensitive to viewpoint selection and requiring repeated per-query reasoning, which limits efficiency and reusability. Consequently, 3D visual grounding remains challenging due to limitations in spatial reasoning, compositional language understanding, and generalization to unseen environments.

\section{Preliminaries and Problem Definition}
\label{sec:prelims}


\subsection{Problem Statement}

3D visual grounding aims to localize a target object in a 3D scene given a natural language query. Let $\mathcal{I} = \{I_1, I_2, \dots, I_n\}$ denote a sequence of RGB-D images capturing a 3D environment, where each $I_i = (I^{RGB}_i, I^D_i)$ consists of a color image and a corresponding depth map. Using these observations, the scene is reconstructed into a set of 3D objects with geometric representations. Let $\mathcal{O} = \{o_1, o_2, \dots, o_m\}$ denote the set of reconstructed object instances in the scene.

We represent the scene as a 3D scene graph $\mathcal{G}_s = (\mathcal{V}_s, \mathcal{E}_s)$, where:
\begin{itemize}
    \item $\mathcal{V}_s$ corresponds to object nodes, each associated with geometric and semantic attributes;
    \item $\mathcal{E}_s$ represents spatial or semantic relationships between scene object pairs.
\end{itemize}

Given a natural language query $q$, the goal of 3D visual grounding is to identify the target object $o^* \in \mathcal{O}$ that best matches the query description. The query may involve a target object and one or more landmark objects, along with object attributes and relational constraints connecting multiple objects.

Formally, grounding can be viewed as a structured matching problem between the linguistic structure of $q$ and the relational structure encoded in $\mathcal{G}_s$. The objective is to select:

\[
o^* = \argmax_{o_i \in \mathcal{O}} \; \mathcal{S}(o_i, q, \mathcal{G}_s),
\]

where $\mathcal{S}(\cdot)$ measures the compatibility between a candidate object and the query under the relational constraints defined by the scene graph.

\section{Method}
\label{sec:method}

\subsection{3D Scene Graph Reconstruction}

We reconstruct the 3D scene from RGB-D sequences using an object-centric approach inspired by \cite{gu2023conceptgraphsopenvocabulary3dscene}, and introduce marker-guided relationship extraction to enable efficient scene graph generation.

\subsubsection{Scene Reconstruction}  
For each RGB-D frame, we first run a class-agnostic object detector to obtain coarse bounding boxes. Each bounding box is then processed by a segmentation model (e.g., SAM\cite{kirillov2023segment}) to generate precise object masks. The pixels within each mask are back-projected into 3D using the corresponding depth map and camera pose, forming partial object-level point clouds. 
To maintain consistent 3D object instances across frames, newly detected objects are matched to existing ones based on geometric overlap and semantic similarity. When a match is found, the new detection is merged into an existing object; otherwise, the new detection is initialized as a new object. 

\subsubsection{Marker-Guided Relationship Extraction}  
Unlike ConceptGraphs~\cite{gu2023conceptgraphsopenvocabulary3dscene}, which predicts relationships by separately querying an LLM with object pairs and their coordinates, we adopt a visual marker prompting strategy~\cite{yang2023setofmarkpromptingunleashesextraordinary}. Each object mask is annotated with a unique marker (numeric or alphabetic) placed at its center. The annotated image is then fed into a VLM, which generates object descriptions and spatial relationships (e.g., ``cup[1] is on top of table[3]''). Because the VLM refers to the object not only by its name but also by its unique marker, relationships can be unambiguously associated with their corresponding object masks.

\subsubsection{2D-to-3D Graph Construction}  
Given that each 2D segmentation mask corresponds to a reconstructed 3D object via RGB-D back-projection, the predicted 2D relationships can be directly lifted into 3D. This produces a scene graph
\[
\mathcal{G}_s = (\mathcal{V}_s, \mathcal{E}_s),
\]

where nodes $\mathcal{V}_s$ represent reconstructed objects, and edges $\mathcal{E}_s$ encode spatial relationships predicted by the VLM.

This design enables joint 3D reconstruction and relationship extraction using 2D visual context, rather than querying language models over pairs of objects defined only by 3D coordinates. This leads to more coherent scene graphs, resulting in improved downstream visual grounding performance.

\subsection{Visual Grounding via Graph Matching}
Once the 3D scene graph $\mathcal{G}_s$ is constructed, we perform visual grounding by matching a query graph $\mathcal{G}_q$ to the scene graph. The query graph $\mathcal{G}_q$ is derived from the referring expression via an LLM and consists of nodes representing objects mentioned in the expression and edges representing relations between them (e.g., ``left of,'' ``behind''). We set the first node to be the target object of the query to maintain a consistent representation across queries. The goal is to find a subgraph in the 3D scene graph that best matches the query graph, where each query graph node maps to a scene node that correctly identifies the object while respecting relational constraints between query graph nodes. 

To reduce the search space, we first perform semantic filtering for each query node $q \in \mathcal{V}_q$. We employ a pre-trained text encoder $\Phi_{txt}$ to compute the embedding $\mathbf{e}_q = \Phi_{txt}(\text{label}(q))$. We then identify the best-matching semantic class $L_s$ within the label space $\mathcal{L}$ by maximizing cosine similarity:
\begin{equation}
    L_s = \argmax_{l \in \mathcal{L}} \, \cos(\mathbf{e}_q, \Phi_{txt}(l)).
\end{equation}
A candidate set $C_q$ is formed by collecting all scene nodes $v_s \in \mathcal{G}_s$ whose labels match $L_s$. 


    

With candidate sets established, we enumerate mappings by fixing the target node $q_0$ to a candidate scene object and searching for assignments of the remaining query nodes. The assignment is performed via depth-first search (DFS) over candidate sets, with the option to skip landmark nodes when necessary. During the search, we maintain the current mapping and track the used scene nodes to avoid duplicate assignments. We represent each mapping $\mathcal{M}$ with an assignment function $m:\mathcal{V}_q \rightarrow \mathcal{V}_s$, where $m(q_i)$ is the scene node assigned to query node $q_i$. Each completed mapping is evaluated using a scoring function that combines node and edge similarities, with an emphasis on correct target grounding:
\begin{equation}
\label{eq:graph_score}
\begin{aligned}
S(\mathcal{M}) &= \alpha S_t\!\left(q_0, m(q_0)\right) + \beta \bar{S}_n + \gamma \bar{S}_e + \delta C(\mathcal{M}) \\
\bar{S}_n &= \frac{1}{|\mathcal{M}|} \sum_{q_i \in \mathcal{V}_q} S_n\!\left(q_i, m(q_i)\right) \\
\bar{S}_e &= \frac{1}{|\mathcal{E}_q|} \sum_{(q_i, q_j) \in \mathcal{E}_q} S_e\!\left((m(q_i), m(q_j)), (q_i, q_j)\right) \\
C(\mathcal{M}) &= \frac{|\mathcal{M}|}{|\mathcal{V}_q|}
\end{aligned}
\end{equation}

\begin{figure*}[t] 
    \centering
    \includegraphics[width=0.85\textwidth]{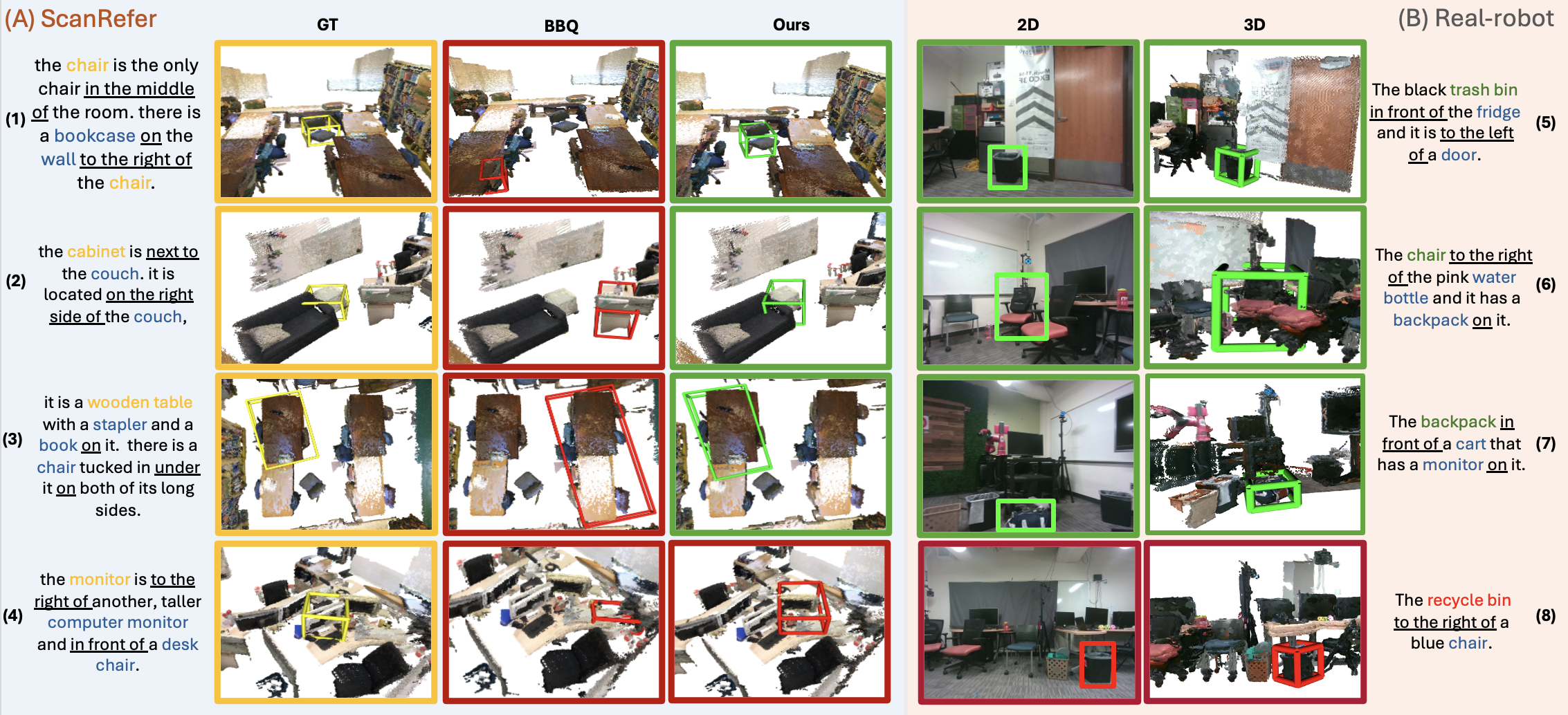}
    \caption{Qualitative results. Left: ScanRefer. Right: real-world robot experiments. Rendered images highlight ground truth objects (\textcolor{yellow}{yellow}), correctly identified objects (\textcolor{green}{green}), and incorrectly identified objects (\textcolor{red}{red}).}
    \label{fig:combo_examples}
\end{figure*}
The $S_t\!\left(q_0, m(q_0)\right)$ and $S_n\!\left(q_i, m(q_i)\right)$ terms measure semantic similarity between each query node and its assigned scene node, computed using CLIP-based semantic similarity or a VLM-based evaluator. Edge similarity $S_e$ compares each query relation $(q_i, q_j) \in \mathcal{E}_q$ with relations between the corresponding assigned scene nodes $(m(q_i), m(q_j))$ in $\mathcal{G}_s$, using CLIP-based text alignment. The final score $S(\mathcal{M})$ also includes a completion term $C(\mathcal{M})$ measuring the fraction of query nodes successfully mapped. This scoring encourages correct identification of the target $q_0$, consistency with other nodes $q \in \mathcal{V}_q$, and alignment of relational edges $\mathcal{E}_q$.


This approach leverages the graph structure explicitly: query edges $\mathcal{E}_q$ are compared with scene edges $\mathcal{E}_s$ during scoring, so mappings $\mathcal{M}$ that preserve relational consistency are favored. By pruning candidates based on labels and similarity, fixing the target $q_0$, and systematically exploring landmark assignments, the method efficiently identifies the target object in 3D scenes while respecting the compositional and relational cues described in the natural language query.

\subsection{Perspective-View Tie-Break via Vision–Language Model}

In scenarios where graph-based grounding produces ambiguous or weak predictions, we employ a VLM to confirm or override the top candidate. Specifically, a perspective RGB image of the 3D scene is rendered and sent to the VLM to select the best target object.

\subsubsection{VLM Gate}  
The VLM is invoked only under conditions where the graph matcher alone is insufficient to resolve ambiguity. These conditions include:  

\begin{itemize}
\item \textbf{Close target candidates with neighbor overlap}: If there are at least two target candidates from graph matching and they share sufficient one-hop neighbors in the scene graph and are spatially close to each other.
\item \textbf{No landmarks or room only}: When the query graph contains only the target node, providing no relational structure for disambiguation, or when the target is described solely in terms of its relative location within the room.
\end{itemize}
\subsubsection{Vision-Language Input for Perspective Disambiguation}
We render a perspective RGB view of the reconstructed 3D scene, focusing on candidate objects identified during graph matching. These objects are highlighted and assigned unique identifiers, ensuring that each visual instance can be unambiguously referenced. On top of the image, we also feed the textual inputs that provide a structured description of the scene and task, including: (i) the free-form open-vocabulary target object query provided by the human, (ii) objects including object labels, unique markers, and object captions.

The VLM is prompted to choose the best candidate among multiple plausible targets. To ensure compatibility with downstream reasoning, the VLM is required to produce a discrete output corresponding to a candidate identifier or a null prediction. The visual and textual inputs are explicitly aligned: all objects referenced in the text correspond to a bounding box in the rendered image. This consistent grounding enables the VLM to jointly reason over appearance, semantics, and spatial relationships. 

This formulation allows the VLM to complement graph-based matching by leveraging holistic visual context, improving robustness in cases involving ambiguity, weak relational cues, or incomplete scene structure.

\section{Experiments}
\label{sec:experiments}

\begin{table*}[t]
\centering
\small
\setlength{\tabcolsep}{4pt}
\renewcommand{\arraystretch}{1.1}
\resizebox{0.8\textwidth}{!}{
\begin{tabular}{l|c|c|c||cc|cc|cc}
\hline
\textbf{Method} & \textbf{\shortstack{GT Pointcloud}} & \textbf{Supervision} & \textbf{Agent} 
& \multicolumn{2}{c|}{\textbf{Unique}} 
& \multicolumn{2}{c|}{\textbf{Multiple}} 
& \multicolumn{2}{c}{\textbf{Overall}} \\
\cline{5-10}
& & & 
& Acc@0.25 & Acc@0.5 
& Acc@0.25 & Acc@0.5 
& Acc@0.25 & Acc@0.5 \\
\hline
\hline

ScanRefer\cite{chen2020scanrefer3dobjectlocalization} & YES & Full & - 
& 67.6 & 46.2 & 32.1 & 21.3 & 39.0 & 26.1 \\

InstanceRefer\cite{yuan2021instancerefer} & YES & Full & - 
& 77.5 & 66.8 & 31.3 & 24.8 & 40.2 & 32.9 \\

3DVG-T\cite{zhao2021_3DVG_Transformer}& YES & Full & - 
& 77.2 & 58.5 & 38.4 & 28.7 & 45.9 & 34.5 \\

BUTD-DETR\cite{jain2022detectiontransformerslanguagegrounding}& YES & Full & - 
& 84.2 & 66.3 & 46.6 & 35.1 & 52.2 & 39.8 \\

EDA\cite{Wu_2023} & YES & Full & - 
& 85.8 & 68.6 & 49.1 & 37.6 & 54.6 & 42.3 \\

3D-VisTA\cite{3dvista} & YES & Full & - 
& 81.6 & 75.1 & 43.7 & 39.1 & 50.6 & 45.8 \\

G3-LQ\cite{wang2024g} & YES & Full & - 
& 88.6 & 73.3 & 50.2 & 39.7 & 56.0 & 44.7 \\

MCLN\cite{qian2024multi} & YES & Full & - 
& 86.9 & 72.7 & 52.0 & 40.8 & 57.2 & 45.7 \\

\hline

WS-3DVG\cite{wang2023distilling} & YES & Weak & - 
& - & - & - & - & 27.4 & 22.0 \\

OpenScene\cite{Peng2023OpenScene} & YES & Zero-Shot & CLIP 
& 20.1 & 13.1 & 11.1 & 4.4 & 13.2 & 6.5 \\

ZSVG3D\cite{yuan2024visualprogrammingzeroshotopenvocabulary} & YES & Zero-Shot & GPT-4 turbo 
& 63.8 & 58.4 & 27.7 & 24.6 & 36.4 & 32.7 \\

SeeGround\cite{li2025seegroundgroundzeroshotopenvocabulary} & YES & Zero-Shot & Qwen2-VL-72B 
& 75.7 & 68.9 & 34.0 & 30.0 & 44.1 & 39.4 \\

\hline

LERF\cite{kerr2023lerf} & NO& Zero-Shot & CLIP 
& - & - & - & - & 4.8 & 0.9 \\

LLM-G\cite{yang2023llmgrounderopenvocabulary3dvisual} & NO & Zero-Shot & GPT-3.5 
& - & - & - & - & 14.3 & 4.7 \\

LLM-G\cite{yang2023llmgrounderopenvocabulary3dvisual} & NO & Zero-Shot & GPT-4 turbo 
& - & - & - & - & 17.1 & 5.3 \\

BBQ\cite{linok2025barequeriesopenvocabularyobject} & NO & Zero-Shot & GPT4-o & - & - & - & - & 19.4 & 11.6\\

\hline

\textbf{Ours} & NO & Zero-Shot & Qwen2.5-VL-7B & 55.7 &39.4 & 24.7 & 15.4 & 34.3& 22.8 \\

\textbf{Ours} & NO & Zero-Shot & gemma4 & 57.3 & 41.7 & 24.1 & 15.5 & 34.4& 23.6\\

\textbf{Ours} & NO & Zero-Shot & Qwen3-VL-8B & \textbf{65.5} & \textbf{58.0} & 27.3 & \textbf{18.4} & 36.8 & \textbf{26.0}\\

\textbf{Ours} & NO & Zero-Shot & GPT-5.3 & 57.7 & 42.0  & \textbf{30.3} & 17.7 & \textbf{38.7}& 25.2\\

\hline
\end{tabular}
}
\caption{Results on ScanRefer validation set.}
\label{tab:main_results}
\end{table*}


\subsection{Datasets}
We evaluate our proposed 3DVG approach on the ScanRefer \cite{chen2020scanrefer3dobjectlocalization} benchmark. ScanRefer consists of 51,500 natural language descriptions across 800 ScanNet \cite{dai2017scannetrichlyannotated3dreconstructions} scenes, where each query specifies a target object's spatial context. The dataset is categorized into two subsets based on scene complexity:
\begin{itemize}
    \item \textbf{Unique}: The target object is the sole instance of its semantic class within the scene.
    \item \textbf{Multiple}: Multiple instances of the same semantic class are present, requiring disambiguation based on spatial relations and object attributes.
\end{itemize}



Following \cite{yang2023llmgrounderopenvocabulary3dvisual,linok2025barequeriesopenvocabularyobject}, we conduct our evaluation on the first 14 scans of the validation set. We report performance at two commonly used thresholds:
\begin{itemize}
    \item \textbf{Acc@0.25}, which reflects coarse localization performance under a relaxed overlap criterion.
    \item \textbf{Acc@0.5}, which evaluates precise grounding under a stricter localization requirement.
\end{itemize}

\subsection{Implementation Details}

Our framework employs YOLO-v8\cite{yolov8_ultralytics} as an open-vocabulary object detector. The detected bounding boxes are forwarded to the Segment Anything Model (SAM-L) to obtain instance-level segmentation masks for each object.

For object representation and merging, we use CLIP ViT-H/14 as the visual encoder to extract features from batched cropped detections, which are aligned with the corresponding class-name text embeddings. 

Qwen3-VL-Instruct-8B\cite{yang2025qwen3} is used for both captioning and grounding: it generates object-focused and relation-aware 2D captions. For grounding, it parses the query into a structured query graph of objects and relationships, identifying the target, associated landmarks, and their relationships.

Finally, alignment between the query target node and the scene node is evaluated using a VLM. We conduct experiments with four VLM variants, including Qwen3-VL-8B, Qwen2.5-VL-7B, GPT-5.3, and Gemma4. The model outputs a semantic similarity score $S_t \in [0,1]$, where higher values indicate stronger alignment between query and scene semantics. The same VLM is also used for target object selection when the VLM gate is activated.

\begin{figure}[htbp]
    \centering
    \begin{subfigure}[t]{0.56\columnwidth}
        \centering
        \includegraphics[width=\linewidth]{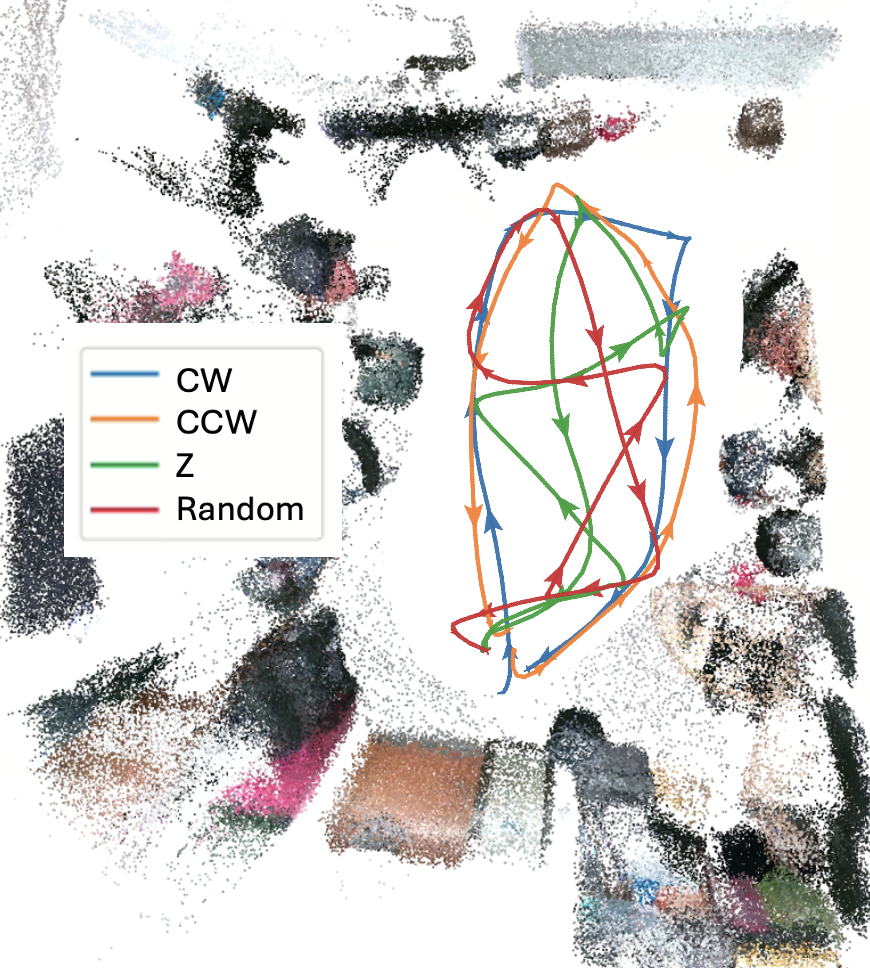}
        \caption{Different trajectory.}
        \label{fig:robot_setup_traj}
    \end{subfigure}
    \hfill
    \begin{subfigure}[t]{0.36\columnwidth}
        \centering
        \includegraphics[width=\linewidth]{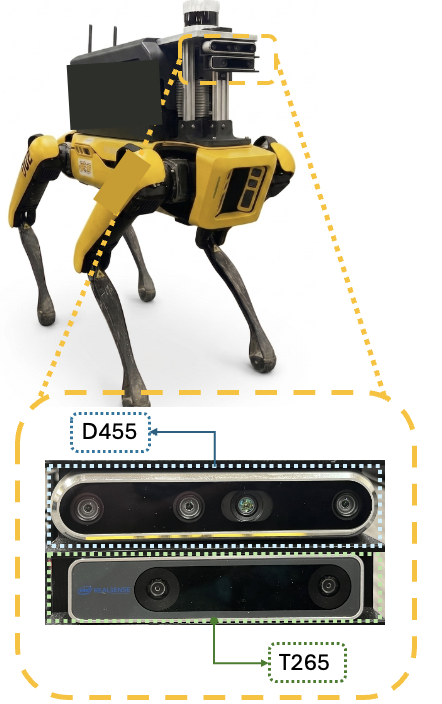}
        \caption{Robot setup.}
        \label{fig:robot_setup_spot}
    \end{subfigure}

    \caption{Robot experiment setup. \textbf{Left}: bird's-eye view of trajectories taken by our quadrupedal robotics platform. \textbf{Right}: the platform and onboard sensing suite used in our experiments.}
    \label{fig:robot_setup_group}
\end{figure}
\subsection{Real-World Deployment}

To validate the practical applicability of our approach, we deploy our system on a Boston Dynamics Spot robot. The platform is equipped with Intel RealSense sensors, including a \textit{D455} RGB-D camera for high-resolution depth perception and a \textit{T265} tracking camera for visual-inertial odometry. This configuration enables reliable scene understanding and real-time pose estimation. All perception and reconstruction components are executed on a remote server equipped with an NVIDIA GeForce RTX 3090.


We evaluate the system in indoor environments within our facilities, where the robot explores the room while collecting RGB-D data and camera odometry. On average, the robot traverses each environment for two full loops to ensure sufficient scene coverage. Quantitative and qualitative results are presented in Section~\ref{sec:result}.

\section{results}
\label{sec:result}

\subsection{Main Results on ScanRefer}
Table~\ref{tab:main_results} presents the performance of our method on the ScanRefer validation set under both unique and multiple object splits. We compare against a wide range of baselines, including fully supervised 3D grounding methods, weakly supervised approaches, and zero-shot methods with and without access to point clouds.

Overall, our method achieves competitive performance among zero-shot approaches while relying solely on RGB-D observations without requiring ground-truth point clouds or precomputed 3D object proposals. Our best-performing variant (Qwen3-VL-8B) attains 36.8\% Acc@0.25 and 26.0\% Acc@0.5 on the ScanRefer validation set, performing on par with prior zero-shot methods such as ZSVG3D and SeeGround, despite not using any 3D inputs. Compared to methods relying on large-scale supervision and privileged 3D inputs (e.g., BUTD-DETR and MCLN), our approach remains competitive in this fully zero-shot setting. Across different backbone models, our framework consistently improves performance, showing that stronger multimodal reasoning benefits 3D grounding under limited geometric cues.

\subsubsection{Unique vs. Multiple Splits}

A consistent trend across all methods is the significant performance gap between the unique and multiple splits. As shown in Table~\ref{tab:main_results}, performance on unique scenes is substantially higher (e.g., up to 65.5\% Acc@0.25 for Qwen3-VL-8B) than on multiple scenes (24.1--30.3\% Acc@0.25 across variants).
This gap indicates that the primary challenge in zero-shot 3D grounding lies in resolving object ambiguity in cluttered environments, rather than single-object recognition. While unique scenes mainly require direct grounding of a clearly specified target, multiple scenes require relational reasoning and disambiguation among competing candidates, which remains a key limitation of current VLM-based systems.
\begin{figure}[t]
\centering
\resizebox{0.95\linewidth}{!}{%
\begin{tikzpicture}
  \pgfplotsset{
    baseaxis/.style={
      tick align=outside,
      major tick length=3.8pt,
      axis line style={plotLine, line width=0.7pt},
      tick style={color=plotLine, line width=0.6pt},
      title style={font=\Large, yshift=-2pt, color=black},
      label style={font=\Large, color=black},
      tick label style={font=\large, color=black},
      grid=major,
      grid style={dashed, color=plotGrid, line width=0.45pt},
    },
    barstyle/.style={
      baseaxis,
      ybar,
      area legend,
      ymin=0, ymax=1.0,
      ytick={0,0.2,0.4,0.6,0.8,1.0},
      bar width=18pt,
      enlarge x limits=0.2,
      legend style={
        draw=none,
        fill=plotLegendFill,
        fill opacity=0.94,
        font=\large,
        text=black,
        cells={anchor=west, xshift=2pt},
        at={(0.02,0.98)},
        anchor=north west
      }
    }
  }

  \begin{groupplot}[
    group style={
        group size=2 by 2, 
        horizontal sep=1.35cm, 
        vertical sep=2.75cm 
    },
    width=0.50\textwidth,
    height=0.49\textwidth
  ]

  \nextgroupplot[
    baseaxis, xbar, xmin=0, xmax=1.0, 
    bar shift=0pt,
    ymin=0.5, ymax=24.5,
    enlarge y limits=false,
    xlabel={Mean IoU},
    ytick={1,...,24},
    yticklabels={piano, step stool, bathroom cabinet, window, tissue box, rug, couch, shelf, trash can, toilet, tv, piano bench, coffee table, microwave, bathtub, nightstand, armchair, laptop, backpack, bookshelf, shower curtain, bed, desk, printer},
    /pgfplots/yticklabel style={font=\fontsize{9.5}{10}\selectfont, inner sep=0.35pt, color=black},
    bar width=6.3pt,
    nodes near coords,
    nodes near coords style={font=\fontsize{8.8}{9.2}\selectfont, color=black, xshift=1.7pt, /pgf/number format/fixed, /pgf/number format/precision=2},
    title={(a) Mean IoU by Category},
    xmajorgrids=true, ymajorgrids=false, y dir=reverse
  ]
    \addplot[fill=colorUnique, draw=none] coordinates {(0.8779,1) (0.8448,2) (0.8228,3)};
    \addplot[fill=colorSky, draw=none] coordinates {(0.5622,4) (0.5135,5) (0.5105,6) (0.5101,7) (0.5080,8) (0.4784,9) (0.4588,10) (0.4555,11) (0.4544,12) (0.4517,13)};
    \addplot[fill=colorOrangeLight, draw=none] coordinates {(0.4399,14) (0.4315,15) (0.4243,16) (0.4120,17) (0.3957,18) (0.3885,19) (0.3814,20) (0.3797,21) (0.3741,22) (0.3577,23) (0.3563,24)};

\nextgroupplot[
    ylabel={Acc (\%)},
    symbolic x coords={IoU@50, IoU@25, IoU@10},
    xtick=data,
    ymin=20, ymax=50,
    ytick={20, 25, 30, 35, 40, 45, 50},
    ymajorgrids=true,
    grid style={line width=0.2pt, draw=gray!15},
    axis lines=left,
    enlarge x limits=0.2,
    title={(b) Model Comparison},
    title style={font=\Large, yshift=-2pt, color=black},
    legend style={at={(0.02,0.98)}, anchor=north west, font=\normalsize, draw=none, fill=none},
    mark size=3.2pt
]

\addplot[color=teal, mark=o, line width=2.8pt, nodes near coords, every node near coord/.append style={font=\scriptsize, anchor=south}] 
    coordinates {(IoU@50, 26.0) (IoU@25, 36.8) (IoU@10, 41.6)};
\addlegendentry{Qwen-3}

\addplot[color=orange, mark=o, line width=2.8pt, nodes near coords, every node near coord/.append style={font=\scriptsize, anchor=south}] 
    coordinates {(IoU@50, 25.2) (IoU@25, 38.8) (IoU@10, 43.5)};
\addlegendentry{GPT-5.3}

\addplot[color=gray!70, mark=o, line width=2.8pt, nodes near coords, every node near coord/.append style={font=\scriptsize, anchor=north, yshift=-3pt}] 
    coordinates {(IoU@50, 23.6) (IoU@25, 34.4) (IoU@10, 38.2)};
\addlegendentry{Gemma-4}
\addplot[color=brown!60, mark=o, line width=2.8pt, nodes near coords, every node near coord/.append style={font=\scriptsize, anchor=north, yshift=-8pt}] 
    coordinates {(IoU@50, 22.8) (IoU@25, 34.3) (IoU@10, 37.5)};
\addlegendentry{Qwen-2.5}

  \nextgroupplot[
    ylabel={Acc (\%)},
    symbolic x coords={IoU@50, IoU@25, IoU@10},
    xtick=data,
    ymin=10, ymax=70,
    ytick={10,20,30,40,50,60,70},
    ymajorgrids=true,
    grid style={line width=0.2pt, draw=gray!15},
    axis lines=left,
    enlarge x limits=0.2,
    title={(c) Split Performance},
    title style={font=\Large, yshift=-2pt, color=black},
    legend style={legend columns=2, at={(0.02,0.98)}, anchor=north west, font=\normalsize, draw=none, fill=none},
    mark size=2.9pt
  ]
    \addplot[color=teal, mark=o, dashed, line width=2.8pt] coordinates {(IoU@50,43.00) (IoU@25,57.98) (IoU@10,65.47)};
    \addplot[color=teal, mark=square*, line width=2.8pt] coordinates {(IoU@50,18.42) (IoU@25,27.34) (IoU@10,30.85)};
    \addplot[color=orange, mark=o, dashed, line width=2.8pt] coordinates {(IoU@50,42.02) (IoU@25,57.65) (IoU@10,65.15)};
    \addplot[color=orange, mark=square*, line width=2.8pt] coordinates {(IoU@50,17.69) (IoU@25,30.26) (IoU@10,33.77)};
    \addplot[color=gray!70, mark=o, dashed, line width=2.8pt] coordinates {(IoU@50,41.69) (IoU@25,57.33) (IoU@10,64.50)};
    \addplot[color=gray!70, mark=square*, line width=2.8pt] coordinates {(IoU@50,15.50) (IoU@25,24.12) (IoU@10,26.46)};
    \legend{Qwen-3-unique, Qwen-3-multiple, GPT-5.3-unique, GPT-5.3-multiple, Gemma-4-unique, Gemma-4-multiple}

  \nextgroupplot[
    ylabel={Acc (\%)},
    symbolic x coords={IoU@50, IoU@25, IoU@10},
    xtick=data,
    ymin=10, ymax=50,
    ytick={10,15,20,25,30,35,40,45,50},
    ymajorgrids=true,
    grid style={line width=0.2pt, draw=gray!15},
    axis lines=left,
    enlarge x limits=0.2,
    title={(d) VLM Usage},
    title style={font=\Large, yshift=-2pt, color=black},
    legend style={at={(0.02,0.98)}, anchor=north west, font=\normalsize, legend columns=2, draw=none, fill=none},
    mark size=2.9pt
  ]
    \addplot[color=teal, mark=diamond*, dashdotted, line width=2.8pt] coordinates {(IoU@50,28.16) (IoU@25,40.35) (IoU@10,45.14)};
    \addplot[color=teal, mark=o, line width=2.8pt] coordinates {(IoU@50,21.19) (IoU@25,28.81) (IoU@10,33.44)};
    \addplot[color=orange, mark=square*, dashed, opacity=0.60, line width=2.8pt] coordinates {(IoU@50,28.16) (IoU@25,40.35) (IoU@10,45.14)};
    \addplot[color=orange, mark=o, line width=2.8pt] coordinates {(IoU@50,19.12) (IoU@25,28.81) (IoU@10,33.44)};
    \addplot[color=gray!70, mark=square*, dashed, line width=2.8pt] coordinates {(IoU@50,26.93) (IoU@25,39.43) (IoU@10,44.20)};
    \addplot[color=gray!70, mark=o, line width=2.8pt] coordinates {(IoU@50,16.61) (IoU@25,23.82) (IoU@10,25.71)};
    \legend{Qwen-3-graph, Qwen-3-vlm, GPT-5.3-graph, GPT-5.3-vlm, Gemma-4-graph, Gemma-4-vlm}

  \end{groupplot}
\end{tikzpicture}
}

\caption{Performance analysis on ScanRefer.}
\begingroup
\captionsetup{font=normalsize}
\endgroup
\label{fig:scanrefer_analysis}
\end{figure}
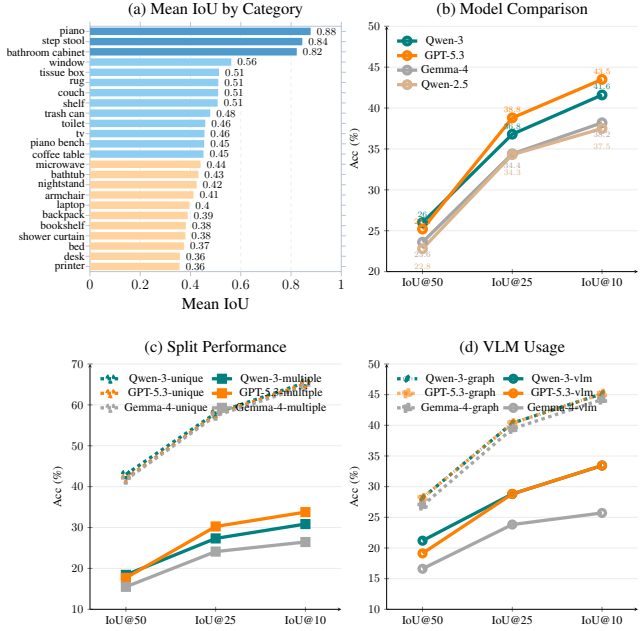
\subsubsection{Graph Matching vs. VLM Selection}

We further analyze the impact of the final selection strategy. As shown in Figure~\ref{fig:scanrefer_analysis}, graph matching achieves higher accuracy than direct VLM selection across all IoU thresholds and model variants. However, this comparison is influenced by a non-uniform invocation mechanism between the two strategies. In our pipeline, VLM selection is typically triggered only when the graph matching module produces low-confidence or ambiguous candidate distributions. As a result, graph matching is more frequently applied to cases with clearly separable candidate objects, whereas VLM is used under more ambiguous conditions with overlapping spatial constraints. This inherent selection bias partially contributes to the observed performance gap. %
Nevertheless, the results highlight the graph matching performs well when relational constraints are sufficiently discriminative, while VLM reasoning complements it by resolving cases that cannot be reliably disambiguated through graph structure alone. More details included in \ref{fig:scanrefer_analysis}.


\begin{table}[htbp]
\centering
\footnotesize
\setlength{\tabcolsep}{2.5pt}
\renewcommand{\arraystretch}{1.05}

\resizebox{0.9\columnwidth}{!}{%
\begin{tabular}{c|ccccc|cc}
  \toprule
  \# & \shortstack{Cand.} & Denoise &
  \shortstack{Graph\\w/o Edges} &
  \shortstack{Graph\\w/ Edges} &
  VLM & Acc@0.5 & Acc@0.25\\
  \midrule
  (a) & \textcolor{green!60!black}{$\checkmark$} & \textcolor{red!75!black}{$\times$} & \textcolor{red!75!black}{$\times$} & \textcolor{red!75!black}{$\times$} & \textcolor{red!75!black}{$\times$} & 4.3 & 12.6\\
  (b) & \textcolor{green!60!black}{$\checkmark$} & \textcolor{green!60!black}{$\checkmark$} & \textcolor{red!75!black}{$\times$} & \textcolor{red!75!black}{$\times$} & \textcolor{red!75!black}{$\times$} & 17.0 & 24.22 \\
  (c) & \textcolor{green!60!black}{$\checkmark$} & \textcolor{green!60!black}{$\checkmark$} & \textcolor{green!60!black}{$\checkmark$} & \textcolor{red!75!black}{$\times$} & \textcolor{red!75!black}{$\times$} & 21.9 & 32.39 \\
  (d) & \textcolor{green!60!black}{$\checkmark$} & \textcolor{green!60!black}{$\checkmark$} & \textcolor{green!60!black}{$\checkmark$} & \textcolor{green!60!black}{$\checkmark$} & \textcolor{red!75!black}{$\times$} & 23.4 & 35.82 \\
  (e) & \textcolor{green!60!black}{$\checkmark$} & \textcolor{green!60!black}{$\checkmark$} & \textcolor{red!75!black}{$\times$} & \textcolor{red!75!black}{$\times$} & \textcolor{green!60!black}{$\checkmark$} & 24.5 & 36.0 \\
  \midrule
  (f) & \textcolor{green!60!black}{$\checkmark$} & \textcolor{green!60!black}{$\checkmark$} & \textcolor{green!60!black}{$\checkmark$} & \textcolor{green!60!black}{$\checkmark$} & \textcolor{green!60!black}{$\checkmark$} & \textbf{26.0} & \textbf{36.8} \\
  \bottomrule
\end{tabular}
}

\caption{Ablation study of different components.}
\label{tab:component_study}
\end{table}

\subsubsection{Effect of Architecture Design}

We evaluate each component of our framework in Table~\ref{tab:component_study} by progressively adding modules on top of the baseline candidate selection.

Starting from candidate selection alone, the model achieves 12.6\% Acc@0.25 and 4.3\% Acc@0.5, indicating that raw proposals are insufficient for reliable grounding.

A key challenge in our setting is the gap between object reconstruction from RGB-D images and methods that rely on ground-truth point clouds or high-quality object proposals. Our reconstructed objects are inherently noisier due to imperfect depth and camera estimation. Adding the denoising module improves performance to 24.2\% Acc@0.25 and 17.0\% Acc@0.5, suggesting that filtering noisy candidates is crucial for stable downstream reasoning.

Graph matching without edges uses the same scoring function as \eqref{eq:graph_score}, except that $\bar{S}_e = 0$ for all node pairs, thereby removing edge-based scoring. The resulting performance drop compared to the graph matching with edges indicates edge information is crucial for mapping the query graph to the scene graph.

With graph matching and VLM selection used independently, both modules yield similar performance improvements, reaching 35.8\%/23.4\% and 36.0\%/24.5\%, respectively. This suggests that each component provides complementary but partially overlapping benefits, with graph matching focusing on spatial relational constraints and VLM contributing semantic disambiguation. When combined, they achieve a substantially stronger result of 36.8\%/26.0\%, indicating that the two sources of information are synergistic in resolving ambiguous candidate objects.

\begin{figure}[htbp]
        \centering
        \resizebox{\columnwidth}{!}{
        \begin{tikzpicture}[font=\sffamily, every node/.append style={text=black}]
    
            \begin{scope}[shift={(-3.05,0)}]
            \begin{scope}[scale=0.72]
                \fill[colorUnique] (0,0) -- (90:2.38) arc (90:201.5:2.38) -- cycle;
                \fill[colorMultiple] (0,0) -- (201.5:2.38) arc (201.5:450:2.38) -- cycle;
                
                \fill[colorVLM] (0,0) -- (90:1.92) arc (90:103.8:1.92) -- cycle;
                \fill[colorNoVLM] (0,0) -- (103.8:1.92) arc (103.8:201.5:1.92) -- cycle;
                \fill[colorVLM] (0,0) -- (201.5:1.92) arc (201.5:297.4:1.92) -- cycle;
                \fill[colorNoVLM] (0,0) -- (297.4:1.92) arc (297.4:450:1.92) -- cycle;
    
                \fill[white] (0,0) circle (1.22);
                
                \node[scale=0.58, align=center, inner sep=0.3pt] at (97:1.54) {VLM\\[-0.12ex]12.4\%};
                \node[scale=0.58, align=center, inner sep=0.3pt] at (152:1.54) {graph\\[-0.12ex]matching\\[-0.12ex]87.6\%};
                \node[scale=0.58, align=center, inner sep=0.3pt] at (250:1.54) {VLM\\[-0.12ex]38.6\%};
                \node[scale=0.58, align=center, inner sep=0.3pt] at (10:1.54) {graph\\[-0.12ex]matching\\[-0.12ex]61.4\%};
    
                \node[align=center, font=\fontsize{6.5}{7}\selectfont] at (145:2.72) {Unique Split\\(31.0\%)};
                \node[align=center, font=\fontsize{6.5}{7}\selectfont] at (-45:2.72) {Multiple Split\\(69.0\%)};
                
                \node[below=2.38cm, font=\bfseries\footnotesize] at (0,0) {(a) VLM / Uniqueness};
    
                \begin{scope}[shift={(2.92, 0.66)}]
                    \foreach \c/\t [count=\i] in {
                        colorUnique/Unique Split,
                        colorMultiple/Multiple Split,
                        colorVLM/VLM Selection,
                        colorNoVLM/{graph matching}
                    } {
                        \fill[\c] (0,-\i*0.31) rectangle (0.22,-\i*0.31+0.155);
                        \node[anchor=west, inner sep=0pt, text=black,
                            font=\sffamily\fontsize{6}{6.5}\selectfont]
                            at (0.24, {-\i*0.31+0.0775}) {\t};
                    }
                \end{scope}
            \end{scope}
            \end{scope}
    
            \begin{scope}[shift={(3.05,0)}]
            \begin{scope}[scale=0.72]
                \foreach \angle/\col/\perc/\num [remember=\angle as \last (initially 90)] in {
                    169.56/fail1/22.1\%/50, 245.88/fail2/21.2\%/48, 
                    308.16/fail3/17.3\%/39, 368.64/fail4/16.8\%/38, 
                    397.44/fail5/8.0\%/18, 418.32/fail6/5.8\%/13, 
                    434.16/fail7/4.4\%/10, 448.56/fail8/4.0\%/9, 
                    450.00/fail9/0.4\%/1} {
                    
                    \fill[\col] (0,0) -- (\last:2.08) arc (\last:\angle:2.08) -- cycle;
                    \draw[white, ultra thin] (0,0) -- (\last:2.08);
                    
                    \node[scale=0.5, align=center, inner sep=0.3pt]
                        at ({(\last+\angle)/2}:1.55) {\perc\\[-0.12ex](\num)};
                }
                
                \node[below=2.58cm, font=\bfseries\footnotesize] at (0,0) {(b) Query Graph Failure Reasons};
    
                \begin{scope}[shift={(2.38,1.05)}]
                    \foreach \c/\t [count=\i] in {
                        fail1/extra relation, fail2/rel. wrong, fail3/missing obj, 
                        fail4/missing rel, fail5/target obj wrong, fail6/lmrk rel. wrong, 
                        fail7/extra obj, fail8/lmrk rel. missing, fail9/lmrk obj wrong} {
                        \fill[\c] (0,-\i*0.31) rectangle (0.22,-\i*0.31+0.155);
                        \node[anchor=west, inner sep=0pt, text=black,
                            font=\sffamily\fontsize{6}{6.5}\selectfont]
                            at (0.24, {-\i*0.31+0.0775}) {\t};
                    }
                \end{scope}
            \end{scope}
            \end{scope}
    
        \end{tikzpicture}
        }
        \vspace{0.05em}

        \includegraphics[width=0.95\columnwidth]{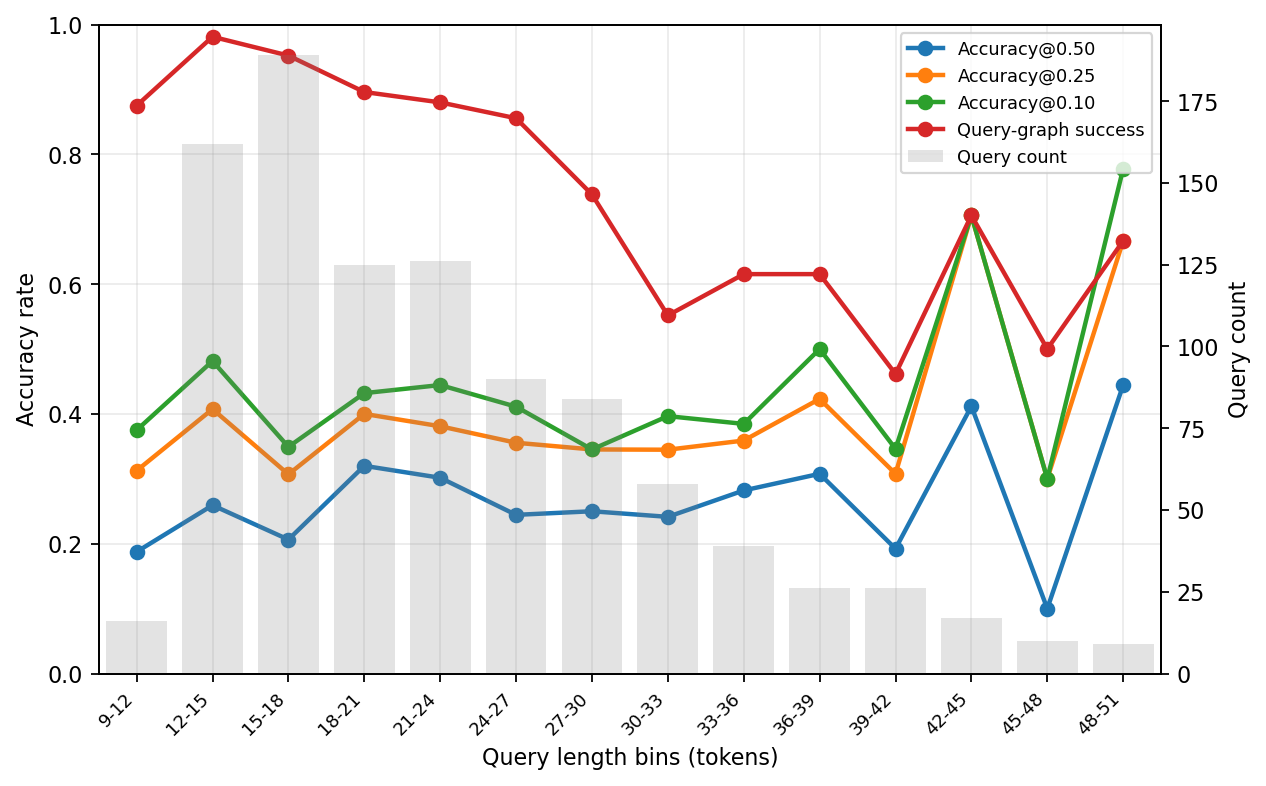}
        
        {\bfseries\footnotesize (c) Caption Length vs. Accuracy}
        \caption{Statistical analysis of the ScanRefer test split: (a) VLM usage vs. unique/multiple query splits; (b) Query parsing failure reason distribution; (c) correlation between query parsing and grounding accuracy.}
        \label{fig:scanrefer_statistical_analysis}
\end{figure}
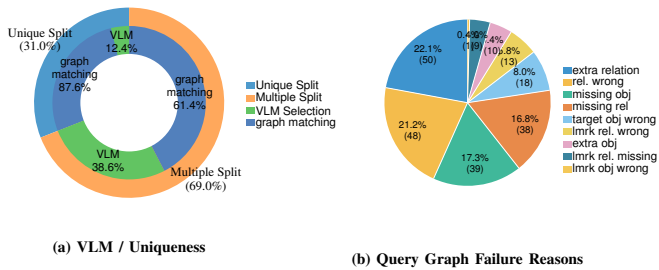
 Compared to
methods relying on large-scale supervision and privileged 3D
inputs (e.g., BUTD-DETR and MCLN), our approach remains
competitive in this fully zero-shot setting. Across different
backbone models, our framework consistently improves per-
formance, showing that stronger multimodal reasoning benefits
3D grounding under limited geometric cue

\subsection{Query Graph Failure Analysis}

We manually inspected and annotated all 991 query graphs to ensure high-quality analysis, achieving an overall accuracy of 825/991 (83.25\%). Figure~\ref{fig:scanrefer_statistical_analysis}(b) provides a breakdown of failure cases when the query graph parsing fails. The majority of errors are attributed to missing or incorrect relational reasoning, including missing relations, incorrect relation interpretation, and failure to identify target objects in multi-object scenes.

We further observe a strong correlation between graph parsing quality and grounding performance on ScanRefer. As shown in Figure~\ref{fig:scanrefer_statistical_analysis}(c), the graph parsing success rate decreases as query length increases, and this trend closely aligns with the degradation in Acc@0.25, Acc@0.5, and Acc@0.10. Longer queries tend to introduce more complex and ambiguous relational structures, which reduces parsing reliability and subsequently impacts final IoU-based grounding performance. 

On the other hand, very short queries do not necessarily yield higher accuracy, as they often lack sufficient relational constraints to uniquely identify the target object, leading to ambiguity among multiple candidates in the scene. This suggests that both under-specification and over-complexity can negatively affect grounding task.


\begin{table}[htbp]
\centering
\caption{Real-Robot Zero-Shot Grounding Experiments.}
\label{tab:robot_summary}
\resizebox{0.9\linewidth}{!}{%
\begin{tabular}{@{}lccccl@{}}
\toprule
\textbf{Target Object} & \textbf{Split Type} & \textbf{Trials} & \textbf{Success} & \textbf{Num of distractors} & \\ \midrule
Chair         & Multiple             & 38               & 18/38              & 6                          \\
Trash bin     & Multiple             & 38               & 17/38              & 4                          \\
Backpack      & Multiple             & 8                & 8/8                & 1                          \\ \midrule
\textbf{Overall} & Multiple          & \textbf{84}      & \textbf{43/84}     & 4.6                        \\ \bottomrule
\end{tabular}%
}
\end{table}

\begin{table}[htbp]
\centering
\caption{The effect of different trajectory.}
\label{tab:dif_traj}
\resizebox{0.6\linewidth}{!}{%
\begin{tabular}{@{}lccccl@{}}
\toprule
\textbf{Loop} & \textbf{Trials} & \textbf{Success} & \textbf{Path Length} \\ \midrule
Clockwise            & 22 & 13/22 & 22.44m \\
Counter Clockwise    & 22 & 8/22  & 27.74m \\
Z-lines              & 22 & 10/22 & 36.95m \\
Random               & 22 & 11/22 & 23.77m \\ \midrule
\textbf{Overall}     & \textbf{88} & \textbf{42/88} & 27.73m \\ \bottomrule
\end{tabular}%
}
\end{table}
  
\subsection{Real-World Robot Experiments}
We evaluate our method in real-world robot experiments, as summarized in Table~\ref{tab:robot_summary}. Overall, we achieve a 51.2\% success rate (43/84), demonstrating effective sim-to-real transfer. Performance varies across object types, with simpler scenes (e.g., backpack) achieving perfect success, while cluttered categories such as chairs and trash bins remain more challenging. {\color{myblue}
Interestingly, we observed that despite containing fewer distractors than the chair category, the trash bin category yields lower performance. This is likely because the scene includes both recycling bins and trash bins, which are differentiated by the language query but not by the reconstruction pipeline. We conclude that objects that are visually similar from the reconstruction perspective but semantically distinct from the query perspective introduce an additional challenge to our framework.} 

\begin{figure}[t]
\centering
\resizebox{\columnwidth}{!}{%
\begin{tikzpicture}
\begin{axis}[
    xbar,
    xmin=0, xmax=30,
    width=0.49\columnwidth,
    height=0.47\columnwidth,
    bar width=5.5pt,
    y dir=reverse,
    symbolic y coords={right of,left of,on,inside,below,in front of,near,between,behind,above,around},
    ytick=data,
    tick label style={font=\fontsize{6.3}{6.8}\selectfont},
    xlabel={Percentage (\%)},
    xlabel style={font=\fontsize{6}{7.4}\selectfont},
    title={(a) Relationship Distribution},
    title style={font=\bfseries\footnotesize},
    xmajorgrids=true,
    grid style={draw=gray!20},
    nodes near coords,
    every node near coord/.append style={font=\fontsize{5.7}{6.2}\selectfont, anchor=west, xshift=1.5pt},
    enlarge y limits=0.03
]
\addplot[fill=colorUnique, draw=none] coordinates {
    (25.35,right of) (19.01,left of) (12.68,on) (10.56,inside) (10.56,below)
    (6.34,in front of) (4.23,near) (3.52,between) (3.52,behind) (2.82,above) (1.41,around)
};
\end{axis}
\end{tikzpicture}
\hspace{0.02\columnwidth}
\begin{tikzpicture}
\begin{axis}[
    xbar,
    xmin=0, xmax=32,
    width=0.49\columnwidth,
    height=0.47\columnwidth,
    bar width=6pt,
    y dir=reverse,
    symbolic y coords={
        VLM verified graph only,
        Small Top-2 Confidence Margin,
        High Target Candidate Ambiguity,
        Incomplete Query Graph Alignment,
        Weak Relational Consistency,
        Low Overall Match Confidence,
        VLM-Based Selection,
        ID failed verifier,
        Target Category Mismatch,
       Low Target-Semantic Alignment
    },
    ytick=data,
    tick label style={font=\fontsize{6.0}{6.5}\selectfont},
    xlabel={Percentage (\%)},
    xlabel style={font=\fontsize{6}{7.4}\selectfont},
    title={(b) Grounding Failure Reasons},
    title style={font=\bfseries\footnotesize},
    xmajorgrids=true,
    grid style={draw=gray!20},
    nodes near coords,
    every node near coord/.append style={font=\fontsize{5.6}{6.1}\selectfont, anchor=west, xshift=1.5pt},
    enlarge y limits=0.03
]
\addplot[fill=fail2, draw=none] coordinates {
    (29.93,VLM verified graph only) (15.33,Small Top-2 Confidence Margin) (14.60,High Target Candidate Ambiguity)
    (12.41,Incomplete Query Graph Alignment) (12.41,Weak Relational Consistency) (5.11,Low Overall Match Confidence)
    (3.65,VLM-Based Selection) (3.65,ID failed verifier) (1.46,Target Category Mismatch) (1.46,Low Target-Semantic Alignment)
};
\end{axis}
\end{tikzpicture}
}
\caption{Real-robot Experiment: (a) relationship-type distribution, and (b) failure reason distribution for wrong grounding.}
\label{fig:robot_pie_placeholder}
\end{figure}
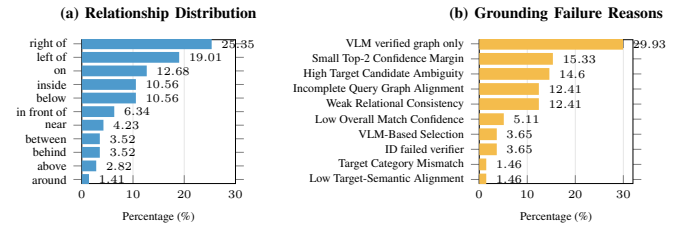

We further study four exploration trajectories: clockwise, counter-clockwise, z-shaped, and random. As shown in Table~\ref{tab:dif_traj}. Overall, the results suggest that grounding performance is less sensitive to the specific trajectory and more dependent on whether objects are observed from sufficient and appropriate viewpoints during exploration. This indicates that effective scene coverage and reasonable scanning viewpoints play a more important role than the exact traversal path. Failure analysis over all 88 trials at Fig.~\ref{fig:robot_pie_placeholder}.

\section{Conclusion and Future Work}
\label{conclusuon}
{
\color{myblue}
We present SceneGraphGrounder, a framework that formulates 3D visual grounding as structured graph matching between a query graph and a 3D scene graph. Our method builds an object-centric scene representation from RGB-D inputs and introduces a marker-guided relationship extraction strategy for efficient and unambiguous relation construction. Grounding is formulated as graph matching between a query graph and the scene graph, enabling joint reasoning over object semantics and multi-hop spatial relations. For ambiguous cases, we introduce a perspective-view tie-break module that leverages a vision–language model as a complementary component. Experiments on ScanRefer and real-world deployments on a Boston Dynamics Spot platform show that explicit structural representations improve grounding accuracy and robustness in open-vocabulary settings.

Currently, our framework assumes a static scene representation; however, real-world robotic applications often demand operation in dynamic environments where object positions and relationships evolve over time. Future research will focus on bridging the gap between static semantic mapping and real-world dynamic interaction remains a critical frontier for autonomous spatial reasoning systems.

}

\bibliographystyle{IEEEtran} 
\bibliography{main}

@misc{wu2023incremental3dsemanticscene,
      title={Incremental 3D Semantic Scene Graph Prediction from RGB Sequences}, 
      author={Shun-Cheng Wu and Keisuke Tateno and Nassir Navab and Federico Tombari},
      year={2023},
      eprint={2305.02743},
      archivePrefix={arXiv},
      primaryClass={cs.CV},
      url={https://arxiv.org/abs/2305.02743}, 
}

@misc{wald2020learning3dsemanticscene,
      title={Learning 3D Semantic Scene Graphs from 3D Indoor Reconstructions}, 
      author={Johanna Wald and Helisa Dhamo and Nassir Navab and Federico Tombari},
      year={2020},
      eprint={2004.03967},
      archivePrefix={arXiv},
      primaryClass={cs.CV},
      url={https://arxiv.org/abs/2004.03967}, 
}

@misc{wu2021scenegraphfusionincremental3dscene,
      title={SceneGraphFusion: Incremental 3D Scene Graph Prediction from RGB-D Sequences}, 
      author={Shun-Cheng Wu and Johanna Wald and Keisuke Tateno and Nassir Navab and Federico Tombari},
      year={2021},
      eprint={2103.14898},
      archivePrefix={arXiv},
      primaryClass={cs.CV},
      url={https://arxiv.org/abs/2103.14898}, 
}

@misc{wang2023vlsatvisuallinguisticsemanticsassisted,
      title={VL-SAT: Visual-Linguistic Semantics Assisted Training for 3D Semantic Scene Graph Prediction in Point Cloud}, 
      author={Ziqin Wang and Bowen Cheng and Lichen Zhao and Dong Xu and Yang Tang and Lu Sheng},
      year={2023},
      eprint={2303.14408},
      archivePrefix={arXiv},
      primaryClass={cs.CV},
      url={https://arxiv.org/abs/2303.14408}, 
}

@misc{gu2023conceptgraphsopenvocabulary3dscene,
      title={ConceptGraphs: Open-Vocabulary 3D Scene Graphs for Perception and Planning}, 
      author={Qiao Gu and Alihusein Kuwajerwala and Sacha Morin and Krishna Murthy Jatavallabhula and Bipasha Sen and Aditya Agarwal and Corban Rivera and William Paul and Kirsty Ellis and Rama Chellappa and Chuang Gan and Celso Miguel de Melo and Joshua B. Tenenbaum and Antonio Torralba and Florian Shkurti and Liam Paull},
      year={2023},
      eprint={2309.16650},
      archivePrefix={arXiv},
      primaryClass={cs.RO},
      url={https://arxiv.org/abs/2309.16650}, 
}

@inproceedings{yuan2021instancerefer,
  title={Instancerefer: Cooperative holistic understanding for visual grounding on point clouds through instance multi-level contextual referring},
  author={Yuan, Zhihao and Yan, Xu and Liao, Yinghong and Zhang, Ruimao and Wang, Sheng and Li, Zhen and Cui, Shuguang},
  booktitle={Proceedings of the IEEE/CVF International Conference on Computer Vision},
  pages={1791--1800},
  year={2021}
}

@article{yang2024exploiting,
  title={Exploiting contextual objects and relations for 3d visual grounding},
  author={Yang, Li and Zhang, Ziqi and Qi, Zhongang and Xu, Yan and Liu, Wei and Shan, Ying and Li, Bing and Yang, Weiping and Li, Peng and Wang, Yan and others},
  journal={Advances in Neural Information Processing Systems},
  volume={36},
  year={2024}
}

@misc{chen2020scanrefer3dobjectlocalization,
      title={ScanRefer: 3D Object Localization in RGB-D Scans using Natural Language}, 
      author={Dave Zhenyu Chen and Angel X. Chang and Matthias Nießner},
      year={2020},
      eprint={1912.08830},
      archivePrefix={arXiv},
      primaryClass={cs.CV},
      url={https://arxiv.org/abs/1912.08830}, 
}

@inproceedings{achlioptas2020referit3d,
  title={Referit3d: Neural listeners for fine-grained 3d object identification in real-world scenes},
  author={Achlioptas, Panos and Abdelreheem, Ahmed and Xia, Fei and Elhoseiny, Mohamed and Guibas, Leonidas},
  booktitle={Computer Vision--ECCV 2020: 16th European Conference, Glasgow, UK, August 23--28, 2020, Proceedings, Part I 16},
  pages={422--440},
  year={2020},
  organization={Springer}
}

@misc{yang2023llmgrounderopenvocabulary3dvisual,
      title={LLM-Grounder: Open-Vocabulary 3D Visual Grounding with Large Language Model as an Agent}, 
      author={Jianing Yang and Xuweiyi Chen and Shengyi Qian and Nikhil Madaan and Madhavan Iyengar and David F. Fouhey and Joyce Chai},
      year={2023},
      eprint={2309.12311},
      archivePrefix={arXiv},
      primaryClass={cs.CV},
      url={https://arxiv.org/abs/2309.12311}, 
}

@misc{xu2024vlmgroundervlmagentzeroshot,
      title={VLM-Grounder: A VLM Agent for Zero-Shot 3D Visual Grounding}, 
      author={Runsen Xu and Zhiwei Huang and Tai Wang and Yilun Chen and Jiangmiao Pang and Dahua Lin},
      year={2024},
      eprint={2410.13860},
      archivePrefix={arXiv},
      primaryClass={cs.CV},
      url={https://arxiv.org/abs/2410.13860}, 
}

@misc{li2025seegroundgroundzeroshotopenvocabulary,
      title={SeeGround: See and Ground for Zero-Shot Open-Vocabulary 3D Visual Grounding}, 
      author={Rong Li and Shijie Li and Lingdong Kong and Xulei Yang and Junwei Liang},
      year={2025},
      eprint={2412.04383},
      archivePrefix={arXiv},
      primaryClass={cs.CV},
      url={https://arxiv.org/abs/2412.04383}, 
}

@misc{yuan2024visualprogrammingzeroshotopenvocabulary,
      title={Visual Programming for Zero-shot Open-Vocabulary 3D Visual Grounding}, 
      author={Zhihao Yuan and Jinke Ren and Chun-Mei Feng and Hengshuang Zhao and Shuguang Cui and Zhen Li},
      year={2024},
      eprint={2311.15383},
      archivePrefix={arXiv},
      primaryClass={cs.CV},
      url={https://arxiv.org/abs/2311.15383}, 
}

@misc{dai2017scannetrichlyannotated3dreconstructions,
      title={ScanNet: Richly-annotated 3D Reconstructions of Indoor Scenes}, 
      author={Angela Dai and Angel X. Chang and Manolis Savva and Maciej Halber and Thomas Funkhouser and Matthias Nießner},
      year={2017},
      eprint={1702.04405},
      archivePrefix={arXiv},
      primaryClass={cs.CV},
      url={https://arxiv.org/abs/1702.04405}, 
}

@misc{xu2024llavaspacesggvisualinstructtuning,
      title={LLaVA-SpaceSGG: Visual Instruct Tuning for Open-vocabulary Scene Graph Generation with Enhanced Spatial Relations}, 
      author={Mingjie Xu and Mengyang Wu and Yuzhi Zhao and Jason Chun Lok Li and Weifeng Ou},
      year={2024},
      eprint={2412.06322},
      archivePrefix={arXiv},
      primaryClass={cs.CV},
      url={https://arxiv.org/abs/2412.06322}, 
}

@misc{yu2023visuallypromptedlanguagemodelfinegrained,
      title={Visually-Prompted Language Model for Fine-Grained Scene Graph Generation in an Open World}, 
      author={Qifan Yu and Juncheng Li and Yu Wu and Siliang Tang and Wei Ji and Yueting Zhuang},
      year={2023},
      eprint={2303.13233},
      archivePrefix={arXiv},
      primaryClass={cs.CV},
      url={https://arxiv.org/abs/2303.13233}, 
}

@inproceedings{chen2024expanding,
  title={Expanding Scene Graph Boundaries: Fully Open-vocabulary Scene Graph Generation via Visual-Concept Alignment and Retention},
  author={Chen, Zuyao and Wu, Jinlin and Lei, Zhen and Zhang, Zhaoxiang and Chen, Changwen},
  booktitle={European Conference on Computer Vision (ECCV)},
  pages={108--124},
  year={2024}
}

@misc{liu2025relationawarehierarchicalpromptopenvocabulary,
      title={Relation-aware Hierarchical Prompt for Open-vocabulary Scene Graph Generation}, 
      author={Tao Liu and Rongjie Li and Chongyu Wang and Xuming He},
      year={2025},
      eprint={2412.19021},
      archivePrefix={arXiv},
      primaryClass={cs.CV},
      url={https://arxiv.org/abs/2412.19021}, 
}

@misc{chen2024scenegraphgenerationroleplaying,
      title={Scene Graph Generation with Role-Playing Large Language Models}, 
      author={Guikun Chen and Jin Li and Wenguan Wang},
      year={2024},
      eprint={2410.15364},
      archivePrefix={arXiv},
      primaryClass={cs.CV},
      url={https://arxiv.org/abs/2410.15364}, 
}

@misc{zhang2021exploitingedgeorientedreasoning3d,
      title={Exploiting Edge-Oriented Reasoning for 3D Point-based Scene Graph Analysis}, 
      author={Chaoyi Zhang and Jianhui Yu and Yang Song and Weidong Cai},
      year={2021},
      eprint={2103.05558},
      archivePrefix={arXiv},
      primaryClass={cs.CV},
      url={https://arxiv.org/abs/2103.05558}, 
}

@inproceedings{NEURIPS2021_9a555403,
 author = {Zhang, Shoulong and li, shuai and Hao, Aimin and Qin, Hong},
 booktitle = {Advances in Neural Information Processing Systems},
 editor = {M. Ranzato and A. Beygelzimer and Y. Dauphin and P.S. Liang and J. Wortman Vaughan},
 pages = {18620--18632},
 publisher = {Curran Associates, Inc.},
 title = {Knowledge-inspired 3D Scene Graph Prediction in Point Cloud},
 url = {https://proceedings.neurips.cc/paper_files/paper/2021/file/9a555403384fc12f931656dea910e334-Paper.pdf},
 volume = {34},
 year = {2021}
}

@inproceedings{feng20233d,
  title={3d spatial multimodal knowledge accumulation for scene graph prediction in point cloud},
  author={Feng, Mingtao and Hou, Haoran and Zhang, Liang and Wu, Zijie and Guo, Yulan and Mian, Ajmal},
  booktitle={Proceedings of the IEEE/CVF Conference on Computer Vision and Pattern Recognition},
  pages={9182--9191},
  year={2023}
}

@inproceedings{3dvista,
    title={3D-VisTA: Pre-trained Transformer for 3D Vision and Text Alignment},
    author={Ziyu, Zhu and Xiaojian, Ma and Yixin, Chen and Zhidong, Deng and Siyuan, Huang and Qing, Li},
    booktitle={ICCV},
    year={2023}
}

@inproceedings{Peng2023OpenScene,
  title     = {OpenScene: 3D Scene Understanding with Open Vocabularies},
  author    = {Peng, Songyou and Genova, Kyle and Jiang, Chiyu "Max" and Tagliasacchi, Andrea and Pollefeys, Marc and Funkhouser, Thomas},
  booktitle = {Proceedings of the IEEE/CVF Conference on Computer Vision and Pattern Recognition (CVPR)},
  year      = {2023}
}

@misc{linok2025barequeriesopenvocabularyobject,
      title={Beyond Bare Queries: Open-Vocabulary Object Grounding with 3D Scene Graph}, 
      author={Sergey Linok and Tatiana Zemskova and Svetlana Ladanova and Roman Titkov and Dmitry Yudin and Maxim Monastyrny and Aleksei Valenkov},
      year={2025},
      eprint={2406.07113},
      archivePrefix={arXiv},
      primaryClass={cs.CV},
      url={https://arxiv.org/abs/2406.07113}, 
}

@inproceedings{zhao2021_3DVG_Transformer,
    title={{3DVG-Transformer}: Relation modeling for visual grounding on point clouds},
    author={Zhao, Lichen and Cai, Daigang and Sheng, Lu and Xu, Dong},
    booktitle={ICCV},
    pages={2928--2937},
    year={2021}
}

@misc{jain2022detectiontransformerslanguagegrounding,
      title={Bottom Up Top Down Detection Transformers for Language Grounding in Images and Point Clouds}, 
      author={Ayush Jain and Nikolaos Gkanatsios and Ishita Mediratta and Katerina Fragkiadaki},
      year={2022},
      eprint={2112.08879},
      archivePrefix={arXiv},
      primaryClass={cs.CV},
      url={https://arxiv.org/abs/2112.08879}, 
}

@inproceedings{Wu_2023,
   title={EDA: Explicit Text-Decoupling and Dense Alignment for 3D Visual Grounding},
   url={http://dx.doi.org/10.1109/CVPR52729.2023.01843},
   DOI={10.1109/cvpr52729.2023.01843},
   booktitle={2023 IEEE/CVF Conference on Computer Vision and Pattern Recognition (CVPR)},
   publisher={IEEE},
   author={Wu, Yanmin and Cheng, Xinhua and Zhang, Renrui and Cheng, Zesen and Zhang, Jian},
   year={2023},
   month=June, pages={19231–19242} }

@inproceedings{wang2024g,
  title={G\^{} 3-lq: Marrying hyperbolic alignment with explicit semantic-geometric modeling for 3d visual grounding},
  author={Wang, Yuan and Li, Yali and Wang, Shengjin},
  booktitle={Proceedings of the IEEE/CVF Conference on Computer Vision and Pattern Recognition},
  pages={13917--13926},
  year={2024}
}

@inproceedings{qian2024multi,
  title={Multi-branch collaborative learning network for 3d visual grounding},
  author={Qian, Zhipeng and Ma, Yiwei and Lin, Zhekai and Ji, Jiayi and Zheng, Xiawu and Sun, Xiaoshuai and Ji, Rongrong},
  booktitle={European Conference on Computer Vision},
  pages={381--398},
  year={2024},
  organization={Springer}
}

@inproceedings{wang2023distilling,
  title={Distilling coarse-to-fine semantic matching knowledge for weakly supervised 3d visual grounding},
  author={Wang, Zehan and Huang, Haifeng and Zhao, Yang and Li, Linjun and Cheng, Xize and Zhu, Yichen and Yin, Aoxiong and Zhao, Zhou},
  booktitle={Proceedings of the IEEE/CVF International Conference on Computer Vision},
  pages={2662--2671},
  year={2023}
}

@inproceedings{kerr2023lerf,
  title={Lerf: Language embedded radiance fields},
  author={Kerr, Justin and Kim, Chung Min and Goldberg, Ken and Kanazawa, Angjoo and Tancik, Matthew},
  booktitle={Proceedings of the IEEE/CVF international conference on computer vision},
  pages={19729--19739},
  year={2023}
}

@inproceedings{kirillov2023segment,
  title={Segment anything},
  author={Kirillov, Alexander and Mintun, Eric and Ravi, Nikhila and Mao, Hanzi and Rolland, Chloe and Gustafson, Laura and Xiao, Tete and Whitehead, Spencer and Berg, Alexander C and Lo, Wan-Yen and others},
  booktitle={Proceedings of the IEEE/CVF international conference on computer vision},
  pages={4015--4026},
  year={2023}
}

@software{yolov8_ultralytics,
  author = {Glenn Jocher and Ayush Chaurasia and Jing Qiu},
  title = {Ultralytics YOLOv8},
  version = {8.0.0},
  year = {2023},
  url = {https://github.com/ultralytics/ultralytics},
  orcid = {0000-0001-5950-6979, 0000-0002-7603-6750, 0000-0003-3783-7069},
  license = {AGPL-3.0}
}

@inproceedings{bonn2020spatial,
  title={Spatial AMR: Expanded spatial annotation in the context of a grounded Minecraft corpus},
  author={Bonn, Julia and Palmer, Martha and Cai, Zheng and Wright-Bettner, Kristin},
  booktitle={Proceedings of the Twelfth Language Resources and Evaluation Conference},
  pages={4883--4892},
  year={2020}
}

@misc{ray2024taskmotionplanninghierarchical,
      title={Task and Motion Planning in Hierarchical 3D Scene Graphs}, 
      author={Aaron Ray and Christopher Bradley and Luca Carlone and Nicholas Roy},
      year={2024},
      eprint={2403.08094},
      archivePrefix={arXiv},
      primaryClass={cs.RO},
      url={https://arxiv.org/abs/2403.08094}, 
}

@article{paul2016efficient,
  title={Efficient grounding of abstract spatial concepts for natural language interaction with robot manipulators},
  author={Paul, Rohan and Arkin, Jacob and Roy, Nicholas and M Howard, Thomas},
  year={2016},
  publisher={Robotics: Science and Systems Foundation}
}

@misc{radford2021learningtransferablevisualmodels,
      title={Learning Transferable Visual Models From Natural Language Supervision}, 
      author={Alec Radford and Jong Wook Kim and Chris Hallacy and Aditya Ramesh and Gabriel Goh and Sandhini Agarwal and Girish Sastry and Amanda Askell and Pamela Mishkin and Jack Clark and Gretchen Krueger and Ilya Sutskever},
      year={2021},
      eprint={2103.00020},
      archivePrefix={arXiv},
      primaryClass={cs.CV},
      url={https://arxiv.org/abs/2103.00020}, 
}

@misc{yang2023setofmarkpromptingunleashesextraordinary,
      title={Set-of-Mark Prompting Unleashes Extraordinary Visual Grounding in GPT-4V}, 
      author={Jianwei Yang and Hao Zhang and Feng Li and Xueyan Zou and Chunyuan Li and Jianfeng Gao},
      year={2023},
      eprint={2310.11441},
      archivePrefix={arXiv},
      primaryClass={cs.CV},
      url={https://arxiv.org/abs/2310.11441}, 
}

@misc{yang2025qwen3,
  title={Qwen3 Technical Report},
  author={Yang, An and Li, Anfeng and others},
  year={2025},
  eprint={2505.09388},
  archivePrefix={arXiv},
  url={https://arxiv.org/abs/2505.09388}
}

\end{document}